\documentclass[times,twocolumn,final]{elsarticle}

\usepackage{ipcai_arxiv}
\usepackage{framed,multirow}
\usepackage{amssymb}
\usepackage{latexsym}
\usepackage{url}
\usepackage{cite}
\usepackage{amsmath,amssymb,amsfonts,bm}
\usepackage{booktabs}
\usepackage{arydshln}
\usepackage{tabulary}
\usepackage{graphicx,epstopdf}
\usepackage{textcomp}
\usepackage{subcaption}
\usepackage{booktabs}
\usepackage[svgnames,table]{xcolor}
\usepackage{pifont}
\usepackage{arydshln}
\usepackage[colorlinks=true,citecolor=cyan,urlcolor=cyan]{hyperref}
\usepackage{graphicx}
\usepackage{placeins}
\usepackage{fancyhdr}
\usepackage{float}

\begin{document}

\begin{frontmatter}

\title{Text-driven Adaptation of Foundation Models for Few-shot Surgical Workflow Analysis}

\author[3]{Tingxuan Chen\fnref{equal1}}
\ead{tingxuan.chen@tum.de}

\author[1,2,3]{Kun Yuan\fnref{equal1}}
\ead{kun.yuan@ext.ihu-strasbourg.eu}

\author[1,2]{Vinkle Srivastav}
\ead{srivastav@unistra.fr}

\author[3]{Nassir Navab}
\ead{nassir.navab@tum.de}

\author[1,2]{Nicolas Padoy\corref{cor1}}
\ead{npadoy@unistra.fr}

\cortext[cor1]{Corresponding author. Tel.: +33-390413530}

\fntext[equal1]{These authors contributed equally to this work.}

\address[1]{ICube, University of Strasbourg, CNRS, Strasbourg, France}
\address[2]{IHU Strasbourg, Strasbourg, France}
\address[3]{CAMP, Technische Universit\"at M\"unchen, Munich, Germany}

\received{XXX}
\finalform{XXX}
\accepted{XXX}
\availableonline{XXX}
\communicated{XXX}

\begin{abstract}

\textbf{Purpose}: Surgical workflow analysis is crucial for improving surgical efficiency and safety. However, previous studies rely heavily on large-scale annotated datasets, posing challenges in cost, scalability, and reliance on expert annotations. To address this, we propose Surg-FTDA (Few-shot Text-driven Adaptation), designed to handle various surgical workflow analysis tasks with minimal paired image-label data.

\textbf{Methods}: Our approach has two key components. First, Few-shot selection-based modality alignment selects a small subset of images and aligns their embeddings with text embeddings from the downstream task, bridging the modality gap. Second, Text-driven adaptation leverages only text data to train a decoder, eliminating the need for paired image-text data. This decoder is then applied to aligned image embeddings, enabling image-related tasks without explicit image-text pairs.

\textbf{Results}: We evaluate our approach to generative tasks (image captioning) and discriminative tasks (triplet recognition and phase recognition). Results show that Surg-FTDA outperforms baselines and generalizes well across downstream tasks.

\textbf{Conclusion}: We propose a text-driven adaptation approach that mitigates the modality gap and handles multiple downstream tasks in surgical workflow analysis, with minimal reliance on large annotated datasets. {The code and dataset will be released in \url{https://github.com/CAMMA-public/Surg-FTDA}.}
\end{abstract}
\end{frontmatter}

\section{Introduction}\label{sec1}

Surgical workflow analysis is crucial for computer-assisted interventions \citep{maier2022surgical, padoy2019machine}. It requires precise surgical scene understanding and human intent anticipation to provide peripheral assistance feedback to surgeons. Early approaches, due to the lack of sufficient data, rely heavily on feature engineering, incorporating semantic features~\citep{twinanda2016endonet}, surgical scene interaction information~\citep{yuan2022anticipation,murali2023latent}, and other domain-specific knowledge to improve scene understanding. With the rise of deep neural networks and the availability of annotated datasets, recent methods have shifted toward designing architectures that better capture temporal dependencies \citep{czempiel2020tecno,gao2021trans} and other complex features. While these architectures have led to notable improvements, they often exhibit limited generalizability and require large amounts of paired image-label data for training, limiting their scalability. 

\begin{figure*}[t!]
    \centering
    \includegraphics[width=\textwidth]{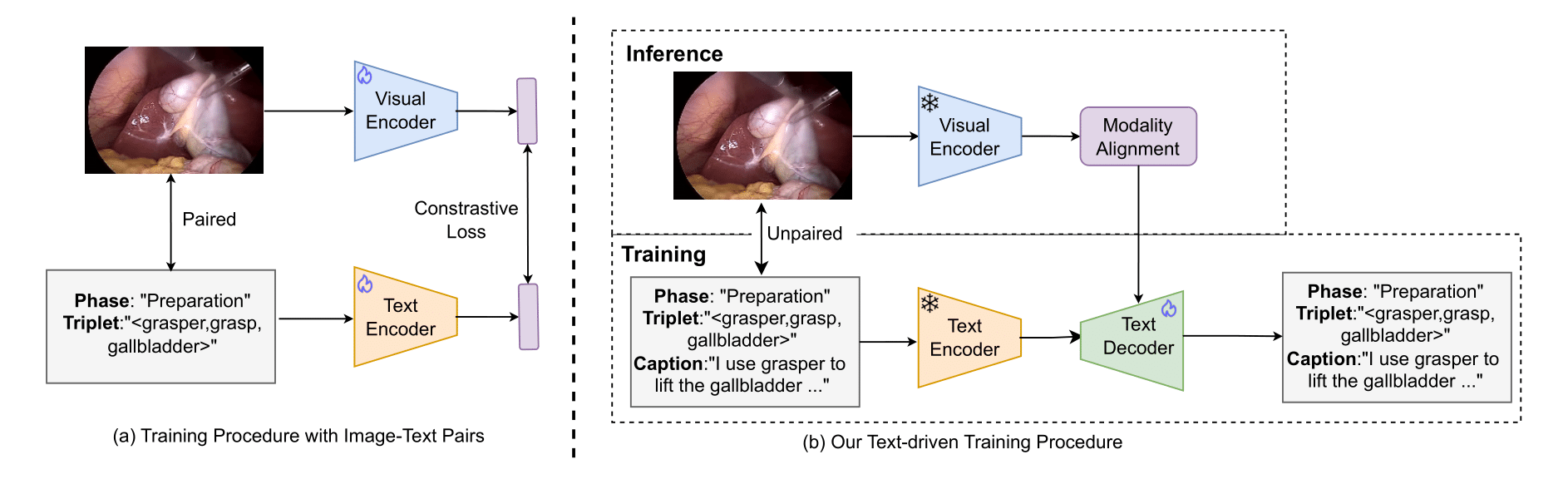} 
    \caption{(a) Conventional adaptation of multi-modal foundation model requires paired image-label data for training; (b) Our text-driven adaptation of the foundation model does not require a large number of image-label pairs to achieve the surgical workflow analysis.}
    \label{fig:fig1}
\end{figure*}

Surgical foundation models are an emerging field leveraging multimodal representation learning to enable systems to interpret visual concepts using natural language \citep{yuan2023learning, yuan2024hecvl, yuan2024procedure}. These models have started to demonstrate good generalizability, allowing adaptation to various surgical procedures and tasks, such as surgical workflow recognition, triplet recognition \citep{nwoye2021rendezvous}, and visual question answering \citep{seenivasan2022surgical,yuan2024advancing}. 

The zero-shot adaptation of these methods, however, still shows a considerable performance gap w.r.t fully-supervised baselines. The main challenge in adapting the surgical multi-modal foundational models to the downstream task is the gap between the vision and the text embeddings on the downstream dataset w.r.t the pre-training dataset. This gap arises when vision and text embeddings are clustered separately, leading to situations where semantically dissimilar images are closer to each other than their corresponding texts. One way to adapt these models to various surgical downstream tasks is through prompt tuning \citep{zhou2022conditional,yao2023visual} or multi-modal fine-tuning \citep{wang2021actionclip} by utilizing the pre-trained visual and textual encoders as illustrated in Fig.~\ref{fig:fig1} (a). However, these adaptation approaches rely heavily on the annotated image-label data pair, which is expensive and limited to specific vocabulary. Moreover, since the pre-trained visual and textual models only consist of encoders, it prevents open-vocabulary generative tasks like image captioning~\citep{russakovsky2015imagenet}.

We address these challenges by proposing a novel adaptation strategy, called Surg-FTDA, to model both discriminative and generative tasks as text-generation problems. Our goal is to transfer a pre-trained surgical multi-modal foundation model to various surgical workflow analysis tasks using minimal paired image-label data. The Surg-FTDA approach operates in two stages. In the first stage, we implement a strategy to select a few data anchors, consisting of image-label pairs from the downstream dataset. This is achieved by performing KMeans clustering on the visual embeddings and selecting data anchors that are maximally distant from each other within the embedding space. These anchors are then used for modality alignment between the visual and textual encoders by projecting both visual and textual embeddings through an MLP layer, followed by minimizing the $L2$ loss between them.

In the second stage, we perform text-only training inspired by CapDec \citep{nukrai-etal-2022-text}, CLOSE\citep{gu2023can}, and SurgVLP\citep{yuan2023learning}. Here, we solely rely on the textual data and use the label text from the downstream dataset for the training. A trainable text decoder is appended to the frozen pre-trained text encoder after the modality alignment from stage one. During training, this operates as an encoder-decoder model, where the frozen text encoder generates textual embeddings from the label text, and the trainable text decoder reconstructs the label texts from these embeddings. During inference, the visual embeddings are generated by the frozen pre-trained visual encoder (after modality alignment from the first stage) and passed through the trained text decoder to generate the corresponding textual labels as the output. Our proposed two-stage training leverages the multi-modal knowledge learned during pre-training and the modality alignment, enabling the effective transfer of visual embeddings to the text decoder for accurate label predictions. We evaluate our approach on both generative tasks, such as image captioning, and discriminative tasks, including triplet recognition and phase recognition. The experimental results demonstrate that Surg-FTDA outperforms baseline methods and exhibits a good generalization across diverse downstream tasks in surgical workflow analysis. We summarize our main contributions as follows:

\begin{itemize} 
    \item We propose a text-driven adaptation approach to transfer surgical multi-modal foundation models to various surgical workflow analysis tasks with minimal paired image-label data. 
    \item We address the modality gap by performing few-shot data selection and modality alignment, which selects a small number of data anchors to learn a transformation matrix between visual and textual modalities. 
    \item We show that our method generalizes well across both discriminative and generative tasks in surgical workflow analysis, reducing the reliance on large-scale annotated datasets. 
\end{itemize}

\section{Methodology}\label{sec2}
In this section, we outline the process of few-shot data anchor selection, text-driven adaptation, and application of our model to both generative and classification tasks.

\begin{figure*}[t!]
    \centering
    \includegraphics[width=\textwidth]{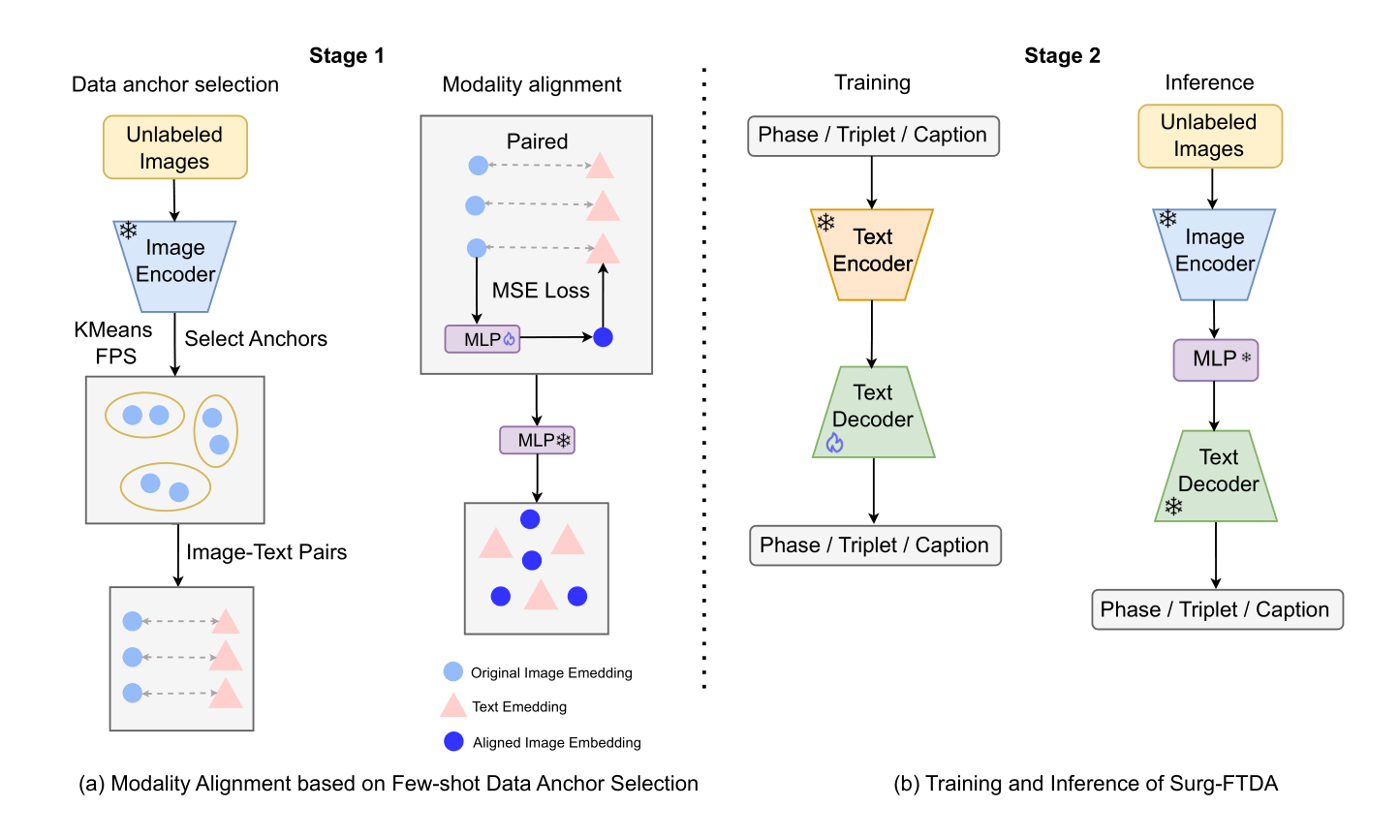}  
    \caption{(a) Few-shot data anchor selection based on the visual embedding space using KMeans or FPS; (b) The text-driven training and inference process of Surg-FTDA, demonstrating how the model applies text-based training to various tasks with minimal paired data.}
    \label{fig:fig2}
\end{figure*}

\subsection{Few-shot Data Anchor Selection}

Few-shot data anchor selection minimizes the need for large-scale annotated image-label pairs during the adaptation of the surgical foundation model. These selected data anchors play a key role in the subsequent text-driven adaptation process, as they help address the modality gap. It is crucial to choose data anchors that are evenly distributed and representative of the unlabeled dataset to ensure effective modality alignment and to enhance the performance of the text-driven adaptation method. 

As shown in Fig.~\ref{fig:fig2} (a), for a downstream dataset with large-scale unlabeled images, we first project these images into the embedding space using the surgical foundation model's image encoder, obtaining the image embedding vectors $V_{\text{image}}^i \in \mathbb{R}^d$, where $i$ represents each image and $d$ is the dimensionality of the embedding space. To identify the data anchors, we apply either KMeans clustering \citep{Hartigan1979} or Farthest Point Sampling (FPS) \citep{qi2017pointnetplusplus}.

For KMeans, the image embeddings are clustered into $K$ clusters, and the embedding vectors closest to the centroids are selected. Alternatively, FPS selects $K$ image embeddings by maximizing the minimum distance between sampled points, ensuring a diverse selection. These methods preserve the structure of the embedding space by selecting a set of diverse data anchors, referred to as $V_{\text{image}}^{\text{sampled}}$. Once the anchor image embeddings are selected, we retrieve their corresponding textual label embeddings, denoted as $V_{\text{text}}^{\text{sampled}}$.

Based on the selected data anchors, we train a Multilayer Perceptron (MLP) to learn a transformation that aligns the image and text modalities. The MLP, denoted as $f_{\text{MLP}}$, takes an image embedding as input and outputs an aligned image embedding. Let $\theta$ represent the parameters of MLP. The MLP is trained by minimizing the MSE loss between the aligned image embeddings $\hat{\mathbf{v}}_{\text{image'}}^{i} = f_{\text{MLP}}(\mathbf{v}_{\text{image}}^{i}; \boldsymbol{\theta})$ and the corresponding text embeddings. The MSE loss is defined as:
\begin{equation}
\frac{1}{K} \sum_{i=1}^{K} \left\lVert \hat{\mathbf{v}}_{\text{image'}}^{i} - \mathbf{v}_{\text{text}}^{i} \right\rVert_2^2
\end{equation} where $K$ is the number of sampled image-label pairs. This process helps reduce the modality gap by aligning the image and text embeddings more effectively.

\subsection{Text-driven Adaptation}
After aligning the vision and language modalities, we propose a text-driven approach to adapt the foundation model for various surgical workflow analysis tasks, treating both discriminative and generative tasks as a text generation problem. We train a text decoder to generate class labels or captions, as depicted in Fig.~\ref{fig:fig2} (b). Specifically, we extract possible target texts from downstream datasets, such as class labels or captions, and fine-tune the decoder to reconstruct these texts based on the extracted text embeddings. 

During the training process, we first extract text embeddings $V_{\text{text}}$ using the frozen text encoder from the foundation model. The decoder is then fine-tuned to reconstruct the original texts from these embeddings:
$\hat{T}^{i} = f_{\text{decoder}}(\mathbf{v}_{\text{text}}^{i}; \boldsymbol{\theta})$, where $f_{\text{decoder}}$ represents the decoder with parameters $\theta$. The learning objective during the fine-tuning is to minimize the auto-regressive cross-entropy loss $\ell$ over all tokens in the target text $T$:
\begin{equation}
    \mathcal{L}_{\text{reconstruction}} = \sum_{T \in \mathcal{T}} \ell(\hat{T}, T)
\end{equation}
where $\hat{T}$ is the predicted text and $T$ is the original text as ground truth from the dataset $\mathcal{T}$.

During inference, we adapt the model to downstream visual tasks by swapping the pre-trained text encoder with the image encoder, allowing the model to accept images as input. We use the frozen image encoder from the foundation model to generate image embeddings $V_{\text{image}}$. However, due to the modality gap between vision and text modalities, we need to align these image embeddings with the text embeddings before feeding them into the decoder. To achieve this, we use the previously trained modality alignment function $f_{\text{MLP}}$ to transform the image embeddings into aligned embeddings, denoted as ${\hat{\mathbf{v}}_{\text{image}'}}$. Then, we pass the aligned image embeddings to the fine-tuned text decoder: $\hat{T} = f_{\text{decoder}}(\hat{V}_{\text{image'}}; \boldsymbol{\theta}),$ where the decoder outputs the task-specific text output, such as class labels or captions. Since the aligned image embeddings are now closer to the text embeddings in the semantic space, the decoder can generate more accurate and contextually relevant text outputs for the task.

\begin{figure*}[]
    \centering
    \includegraphics[width=\textwidth]{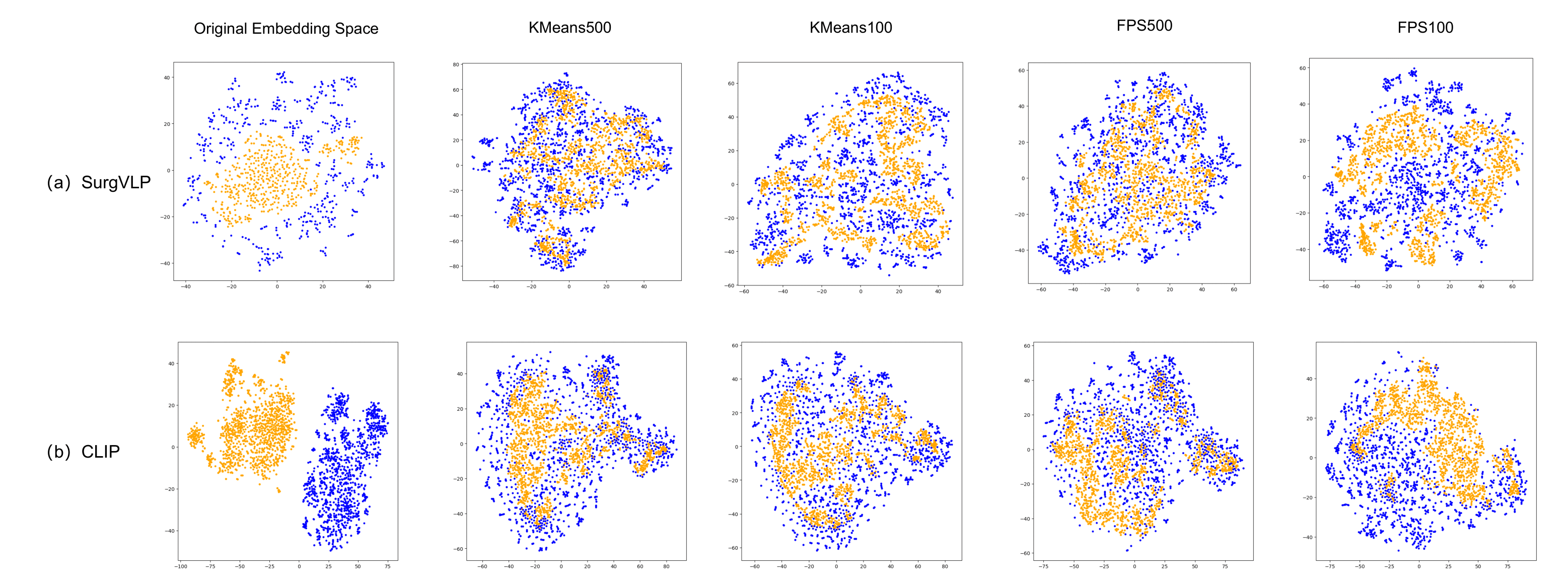}  
    \caption{Visualization of modality gap across foundation models using few-shot data anchor selection and modality alignment. Yellow points represent the image embedding vectors. Blue points represent the text embedding vectors.}
    \label{fig:fig3}
\end{figure*}

\begin{table*}[t!]
\caption{Results of triplet recognition. The numbers in the bracket indicate the number of image-text pairs used by the model.}
\label{tab:tab2}
\centering
\begin{tabular}{lccc ccc}
\toprule
& \multicolumn{3}{c}{SurgVLP} & \multicolumn{3}{c}{CLIP} \\\cmidrule(lr){2-4}\cmidrule(lr){5-7}
Metric & Precision & Recall & F1 & Precision & Recall & F1 \\
\midrule
\multicolumn{7}{c}{\textbf{Fully Supervised Approaches}} \\
Image-Text Training & 33.18 & 30.45 & 31.75 & 29.20 & 31.63 & 30.37 \\
\multicolumn{1}{c}{(8331)} & & & & & &  \\
\midrule
\multicolumn{7}{c}{\textbf{Weakly Supervised Approaches}} \\
Surg-FTDA (500) & \textbf{23.81} & \textbf{23.94} & \textbf{23.87} & \textbf{21.08} & \textbf{19.64} & \textbf{20.33} \\
CapDec (1351) & 20.23 & 21.63 & 20.90 & 10.90 & 16.49 & 13.12 \\
\bottomrule
\end{tabular}
\end{table*}

\begin{table*}
\caption{Results of phase recognition. The numbers in the bracket indicate the number of image-text pairs used by the model.}
\label{tab:tab3}
\setlength{\tabcolsep}{5pt} 
\centering
\begin{tabular}{lcccccccc}
\toprule
& \multicolumn{4}{c}{SurgVLP} & \multicolumn{4}{c}{CLIP} \\\cmidrule(lr){2-5}\cmidrule(lr){6-9}
Metric & ACC & Precision & Recall & F1 & ACC & Precision & Recall & F1 \\
\midrule
\multicolumn{9}{c}{\textbf{Fully Supervised Approaches}} \\
Image-Text Training & 57.79 & 41.35 & 42.27 & 38.37 & 54.04 & 33.61 & 31.65 & 30.37 \\
\multicolumn{1}{c}{(8331)} & & & & & & & & \\
\midrule
\multicolumn{9}{c}{\textbf{Weakly Supervised Approaches}} \\
Surg-FTDA (500) & \textbf{55.33} & \textbf{48.10} & \textbf{37.21} & \textbf{39.02} & \textbf{51.20} & \textbf{27.62} & \textbf{31.59} & \textbf{28.93} \\
CapDec (1351) & 10.46 & 18.31 & 20.77 & 8.92 & 31.63 & 8.64 & 19.92 & 11.01 \\
\bottomrule
\end{tabular}
\end{table*}

\subsection{Multi-task Text Decoder}
In our model, the text decoder behaves differently across different types of tasks. In discriminative tasks, such as triplet recognition and phase recognition, the decoder is trained as a classifier, identifying the decision boundary of the embeddings. In generative tasks, such as image captioning, the decoder functions as a generator, autoregressively producing descriptive text for surgical images.

We also adapt the foundation model to handle multiple tasks jointly by training a single text decoder capable of performing two types of discriminative tasks: triplet recognition and phase recognition. For each image, there are associated labels for both phase and triplet tasks. During training, the input to the model consists of the text embeddings derived from phase labels and triplet labels. The decoder is trained to reconstruct the corresponding textual labels based on these embeddings.

To enable the model to generate task-specific outputs based on tasks (either triplet recognition or phase recognition), we train separate MLP's for image embeddings to phase text embeddings and triplet text embeddings. During inference, the task-specific output is controlled by aligning the image embeddings to the corresponding task embeddings using a task-specific MLP. This alignment ensures that the input to the decoder is tailored to the specific task, allowing the model to generate task-relevant outputs effectively.

\section{Experiments}

In this section, we first visualize the effect of our modality alignment function on reducing the modality gap. Second, we evaluate the model's performance on individual downstream tasks. Third, we conduct ablation studies to assess the contributions of various model components. Last, we test the performance of a multi-task decoder trained with mixed input for different tasks. In these experiments, both the CLIP~\citep{radford2021learning} and SurgVLP~\citep{yuan2023learning} foundation models are used to assess the generalizability of our proposed pipeline. We use GPT-2~\citep{radford2019language} as the text decoder.

The architecture of the MLP for modality alignment consists of an input layer with dimensions matching the input features, two hidden layers with 128 neurons each and ReLU activation, and an output layer with dimensions matching the target features. The model is optimized using the Adam optimizer with a learning rate of 0.001, trained for 15 epochs with a batch size of 16, and the loss function used is Mean Squared Error (MSELoss).

For text-driven adaptation, the GPT-2 decoder~\citep{radford2019language} is trained over 10 epochs using the AdamW optimizer with a learning rate of 2e-5. The batch size is set to 34 by default but can be adjusted based on GPU memory and dataset size.

\subsection{Modality Gap}
{To better demonstrate the effectiveness of our few-shot anchor selection-based modality alignment, we visualize the original distributions of the two modalities and the distributions after modality alignment using various sampling methods on the SVL-Caption validation set, which contains 1351 image-text pairs. These visualizations provide an intuitive understanding of how our alignment approach reduces the modality gap and improves the coherence between the embeddings.}

Fig.~\ref{fig:fig3} demonstrates a significant reduction in the modality gap after aligning vision and text embeddings. Notably, embeddings from different modalities are more closely aligned when using $500$ data anchors compared with using $100$ anchors. Additionally, SurgVLP~\citep{yuan2023learning} provides better initialization for alignment than CLIP~\citep{radford2021learning}, with a smaller initial modality gap. These results confirm that our proposed modality alignment function based on few-shot data anchor selection effectively bridges the modality gap.

\subsection{Downstream Tasks}
In this work, we investigate two discriminative tasks, phase and action triplet recognition, and generative task, image captioning. We compare our text-driven adaptation method with the fully-supervised models that are finetuned on large-scale image-label pairs, and with the weakly supervised CapDec \citep{nukrai-etal-2022-text} approach, which optimizes noise parameters using image-label pairs.

\subsubsection{Discriminative Tasks}
\textbf{Triplet Recognition:} We use triplet text samples from the CholecT50~\citep{nwoye2021rendezvous} dataset to train the decoder through our text-driven adaptation and test the model on images from the test split. The model’s performance is evaluated using precision, recall, and F1 score with macro averaging. As shown in Tab.~\ref{tab:tab2}, our model consistently outperforms CapDec across various metrics.

\textbf{Phase Recognition:} 
We report accuracy, precision, recall, and F1 score on the test split of Cholec80~\citep{twinanda2016endonet} dataset, with macro averaging applied. As shown in Tab.~\ref{tab:tab3}, our model's performance on these metrics closely approaches that of fully-supervised approaches and significantly outperforms CapDec. This demonstrates that our text-driven adaptation trains the text decoder to become a semantic classifier. Additionally, it highlights the capability of the pre-trained foundation model to comprehend both coarse and fine-grained semantics, such as phases and triplets.

\subsubsection{Generative Tasks}
\textbf{Image Caption:} We train the model using caption text in the SVL-Caption dataset. We evaluate the models using standard captioning metrics: BLEU\citep{papineni2002bleu} (B@1, B@4), METEOR\citep{banerjee2005meteor}, and CIDEr\citep{vedantam2015cider}. As shown in Tab.~\ref{tab:tab1}, our model outperforms CapDec\citep{nukrai-etal-2022-text} across various metrics on both foundation models, demonstrating its effectiveness for generative tasks.

\setlength{\tabcolsep}{5pt} 
\begin{table*}
\caption{Results of image caption. The numbers in the bracket indicate the number of image-label pairs used by the model.}
\label{tab:tab1}
\centering
\begin{tabular}{lcccccccc}
\toprule
& \multicolumn{4}{c}{SurgVLP} & \multicolumn{4}{c}{CLIP} \\\cmidrule(lr){2-5}\cmidrule(lr){6-9}
Metric & B@1 & B@4 & METEOR & CIDEr & B@1 & B@4 & METEOR & CIDEr \\
\midrule
\multicolumn{9}{c}{\textbf{Fully Supervised Approaches}} \\
Image-Text Training & 25.76 & 3.21 & 17.41  & 15.02 & 23.23 & 1.92 & 15.59 & 12.84 \\
\multicolumn{1}{c}{(5404)} & & & & & & & & \\
\midrule
\multicolumn{9}{c}{\textbf{Weakly Supervised Approaches}} \\
Surg-FTDA (500) & \textbf{24.60} & \textbf{2.98} & \textbf{18.66} & \textbf{15.65}  & \textbf{25.32} & \textbf{2.43} & \textbf{17.06} & \textbf{11.56} \\
CapDec (1351) & 23.21 & 1.79 & 15.21 & 11.15  & 21.99 & 1.11 & 14.08 & 7.75 \\
\bottomrule
\end{tabular}
\end{table*}

\subsection{Ablation Study}

In this section, we evaluate the impact of our few-shot selection techniques and foundation model choice on both discriminative and generative tasks. We compare KMeans and FPS sampling strategies using $100$ and $500$ sampled anchors, and no-anchor (without modality alignment) on SurgVLP and CLIP models. {We further compare Surg-FTDA with fully-supervised models trained using 10\%, 30\%, 50\%, and 100\% of the image-text pairs from the dataset.}

As shown in Tab.~\ref{tab:ablation_caption}, Tab.~\ref{tab:ablation_triplet}, and Tab.~\ref{tab:ablation_phase}, demonstrate that KMeans generally outperforms FPS across most tasks. This is likely due to KMean's ability to identify more representative data anchors within the dataset, resulting in better alignment between the visual and textual modalities. Also, increasing the number of sampling points improves model's performance by learning a robust modality alignment function. SurgVLP consistently outperforms CLIP, demonstrating the benefit of surgical domain-specific pre-training.

\begin{table*}
\caption{Ablation experiments of image caption. The performance of the model improves with an increasing number of selected data anchors.}
\label{tab:ablation_caption}
\centering
\begin{tabular}{lcccccccc}
\toprule
& \multicolumn{4}{c}{SurgVLP} & \multicolumn{4}{c}{CLIP} \\\cmidrule(lr){2-5}\cmidrule(lr){6-9}
Metric & B@1 & B@4 & METEOR & CIDEr & B@1 & B@4 & METEOR & CIDEr \\
\midrule
No-Anchor & 23.26 & 2.01 & 16.36 & 10.83 & 15.22 & 1.36 & 10.96 & 8.03 \\
KMeans 100 & \textbf{24.81} & 2.84 & 18.64 & 15.35 & 23.62 & 1.34 & 16.33 & 10.00 \\
KMeans 500 & 24.60 & 2.98 & 18.66 & 15.65 & \textbf{25.32} & 2.43 & 17.06 & 11.56 \\
FPS 100 & 22.77 & 2.15 & 16.71 & 10.52 & 24.03 & 1.56 & 15.90 & 7.75 \\
FPS 500 & 24.69 & \textbf{3.26} & \textbf{18.85} & \textbf{15.97} & 24.30 & \textbf{2.64} & \textbf{17.83} & \textbf{12.42} \\
\bottomrule
\end{tabular}
\end{table*}

\begin{table*}
\caption{Ablation experiments of triplet recognition. The performance of the model improves with an increasing number of selected data anchors, and KMeans proves to be more suitable for this task compared to FPS.}
\label{tab:ablation_triplet}
\centering
\begin{tabular}{lcccccccc}
\toprule
& \multicolumn{4}{c}{SurgVLP} & \multicolumn{4}{c}{CLIP} \\\cmidrule(lr){2-5}\cmidrule(lr){6-9}
Metric & Acc & Precision & Recall & F1 & Acc & Precision & Recall & F1 \\
\midrule
No-Anchor & 1.34 & 9.76 & 15.40 & 11.95 & 0.35 & 6.05 & 7.21 & 6.58 \\
KMeans 100 & 3.01 & 13.09 & 17.84 & 15.10 & 19.89 & 14.03 & 8.53 & 10.61 \\
KMeans 500 & \textbf{27.97} & \textbf{23.81} & \textbf{23.94} & \textbf{23.87} & \textbf{26.49} & 21.08 & \textbf{19.64} & \textbf{20.33} \\
FPS 100 & 10.28 & 14.68 & 16.97 & 15.74 & 22.30 & 15.04 & 8.58 & 10.93 \\
FPS 500 & 25.08 & 20.46 & 21.58 & 21.00 & 25.94 & \textbf{23.04} & 15.53 & 18.55 \\
\bottomrule
\end{tabular}
\end{table*}

\begin{table*}
\caption{Ablation experiments of phase recognition. The performance of the model improves with an increasing number of data anchors.}
\label{tab:ablation_phase}
\centering
\begin{tabular}{lcccccccc}
\toprule
& \multicolumn{4}{c}{SurgVLP} & \multicolumn{4}{c}{CLIP} \\\cmidrule(lr){2-5}\cmidrule(lr){6-9}
Metric & ACC & Precision & Recall & F1  & ACC & Precision & Recall & F1  \\
\midrule
No-Anchor & 8.12 & 12.89 & 16.67 & 4.79 & 3.91 & 4.01 & 14.28 & 1.08 \\
KMeans 100 & 50.63 & 29.29 & 26.49 & 24.77 & 44.64 & 13.69 & 17.42 & 14.49 \\
KMeans 500 & \textbf{55.33} & 48.10 & 37.21 & 39.02 & \textbf{51.20} & \textbf{27.62} & 31.59 & \textbf{28.93} \\
FPS 100 & 41.36 & 17.09 & 16.88 & 14.96 & 30.44 & 7.33 & 21.73 & 9.96 \\
FPS 500 & 54.92 & \textbf{48.42} & \textbf{38.58} & \textbf{40.26} & 37.00 & 18.85 & \textbf{32.10} & 21.76 \\
\bottomrule
\end{tabular}
\end{table*}

{As shown in Tab.~\ref{tab:phase_recognition}, for the phase recognition task, our Surg-FTDA significantly outperforms fully supervised models trained with 10\% (833 image-text pairs), 30\% (2499 pairs), and 50\% (4165 pairs) of the dataset. It performs only slightly below the fully supervised model trained on the entire dataset (100\%, 8331 pairs). Remarkably, Surg-FTDA achieves this using only 500 image-text pairs.}

{The fully supervised model trained on 10\% of the data fails to generate phase outputs in the correct format under the same experimental settings (e.g., epochs, learning rate, and hyperparameters), resulting in all metrics being zero. This highlights the limitations of fully supervised models in low-data scenarios and underscores the robust performance of Surg-FTDA even with limited annotated data.}

\begin{table}[h]
\caption{Phase Recognition Performance of Surg-FTDA and Fully Supervised Models}
\label{tab:phase_recognition}
\centering
\begin{tabular}{lcccc}
\toprule
{Metrics} & {ACC} & {Precision} & {Recall} & {F1} \\
\midrule
{10\% (833)} & {0} & {0} & {0} & {0} \\
{30\% (2499)} & {43.37} & {29.29} & {26.49} & {24.77} \\
{50\% (4165)} & {52.47} & {40.66} & {39.48} & {34.44} \\
{100\% (8331)} & {57.79} & {41.35} & {42.27} & {38.37} \\
{Surg-FTDA (500)} & {55.33} & {48.10} & {37.21} & {39.02} \\
\bottomrule
\end{tabular}
\end{table}

{For the triplet recognition task, the results in Tab.~\ref{tab:triplet_recognition} show a similar trend. Surg-FTDA significantly outperforms the fully supervised models trained on 10\% (833 image-text pairs), 30\% (2499 pairs), and 50\% (4165 pairs) of the dataset, while performing slightly below the fully supervised model trained on the full dataset (100\%, 8331 pairs). Similarly, the fully supervised model trained on 10\% of the data still failed to generate valid outputs, resulting in zero metrics, further highlighting the robustness and data efficiency of Surg-FTDA.}

\begin{table}[h]
\caption{{Triplet Recognition Performance of Surg-FTDA and Fully Supervised Models}}\label{tab:triplet_recognition}
\centering
\begin{tabular}{lcccc}
\toprule
{Metrics} & {ACC} & {Precision} & {Recall} & {F1} \\
\midrule
{10\% (833)} & {0} & {0} & {0} & {0} \\
{30\% (2499)} & {12.02} & {17.30} & {22.79} & {20.38} \\
{50\% (4165)} & {17.83} & {15.99} & {20.75} & {18.07} \\
{100\% (8331)} & {30.26} & {33.18} & {30.45} & {31.75} \\
{Surg-FTDA (500)} & {27.97} & {23.81} & {23.94} & {23.87} \\
\bottomrule
\end{tabular}
\end{table}

\subsection{Multi-task Text Decoder for Better Decision Boundary}

In this section, we evaluate the effectiveness of training a single decoder with mixed input for multiple tasks, i.e., phase recognition and triplet recognition. As shown in Tab.~\ref{tab:mixed_triplet} and~\ref{tab:mixed_phase}, the multi-task decoder consistently outperforms task-specific decoders trained on single-task texts, demonstrating that the multi-task learning of text-driven adaptation enhances performance on individual tasks. This suggests that the text decoder learns more robust decision boundaries, enabling it to better distinguish complex semantic features.

\begin{table}[h]
\caption{Results of triplet recognition using multi-task trained decoder. The multi-task trained
decoder performs better than the task-specific trained decoder.}\label{tab:mixed_triplet}
\centering
\begin{tabular*}{\linewidth}{@{\extracolsep{\fill}} lccc}
\toprule
Metric & Precision & Recall & F1  \\
\midrule
Task-Specific Trained & 23.81 & 23.94 & 23.87\\
Multi-Task Trained & \textbf{24.55} & \textbf{24.23} & \textbf{24.39}\\
\bottomrule
\end{tabular*}
\end{table}

\begin{table}[h]
\caption{Results of phase recognition using multi-task trained decoder. The multi-task trained decoder performs better than the task-specific trained decoder.}\label{tab:mixed_phase}
\centering
\begin{tabular*}{\linewidth}{@{\extracolsep{\fill}} lcccc}
\toprule
Metric & ACC & Precision & Recall & F1  \\
\midrule
Task-Specific Trained &55.33& 48.10 & 37.21 & 39.02\\
Multi-Task Trained &\textbf{56.04} & \textbf{48.13} & \textbf{37.79} & \textbf{40.41}\\
\bottomrule
\end{tabular*}
\end{table}

\section{Conclusion}\label{sec13}

In this work, we introduced Surg-FTDA (Few-shot Text-driven Adaptation), a novel approach designed to address the challenges of surgical workflow analysis with minimal reliance on large-scale annotated datasets. By leveraging few-shot data selection and text-driven adaptation, Surg-FTDA bridges the modality gap between vision and text, allowing the model to handle various downstream tasks such as phase recognition, triplet recognition, and image captioning. Our method demonstrates that by selecting a small, diverse subset of image-label pairs and aligning visual and textual embeddings, a text-trained decoder can generalize effectively to visual tasks without requiring large amounts of paired image-label data. The results show that Surg-FTDA outperforms existing baselines, contributing to the field of surgical workflow analysis by enabling more scalable, data-efficient, and versatile models. This few-shot, text-driven adaptation opens up new possibilities for applying foundation models to a broader range of tasks, where limited annotations and multi-modality learning are key challenges.

\section{Acknowledgements}
This work has received funding from the European Union (ERC, CompSURG, 101088553). Views and opinions expressed are however those of the authors only and do not necessarily reflect those of the European Union or the European Research Council. Neither the European Union nor the granting authority can be held responsible for them. This work was also partially supported by French state funds managed by the ANR under Grants ANR-20-CHIA-0029-01 and ANR-10-IAHU-02. This work was granted access to the HPC resources of IDRIS under the allocations AD011013704R1, AD011011631R2, and AD011011631R4 made by GENCI. The authors would like to acknowledge the High-Performance Computing Center of the University of Strasbourg for supporting this work by providing scientific support and access to computing resources. Part of the computing resources were funded by the Equipex Equip@Meso project (Programme Investissements d'Avenir) and the CPER Alsacalcul/Big Data.

\clearpage
\bibliographystyle{model2-names.bst}

\end{document}